%% file: icml2024.tex
\documentclass{article}

\usepackage{microtype}
\usepackage{graphicx}
\usepackage{subfigure}
\usepackage{booktabs} 

\usepackage{hyperref}

\usepackage[accepted]{icml2024}

\usepackage{amsmath}
\usepackage{amssymb}
\usepackage{mathtools}
\usepackage{amsthm}
\usepackage{tabularx}
\usepackage{bbm}

\usepackage[capitalize,noabbrev]{cleveref}

\theoremstyle{plain}

\input{math_commands.tex}

\PassOptionsToPackage{table}{xcolor}
\usepackage{xcolor}

\usepackage{hyperref}
\usepackage{url}

\usepackage{bm}              
\usepackage{booktabs}        
\usepackage{graphicx} 
\definecolor{Gray}{gray}{0.9}
\usepackage{amsmath}
\usepackage{algorithm}
\usepackage{algorithmic}
\usepackage{amsmath}
\usepackage{amssymb}
\colorlet{darkgreen}{green!65!black}
\colorlet{darkblue}{blue!75!black}
\colorlet{darkred}{red!80!black}
\definecolor{lightblue}{HTML}{0071bc}
\definecolor{lightgreen}{HTML}{39b54a}
\definecolor{c1}{HTML}{E83100}
\definecolor{c2}{HTML}{2F70AF}
\definecolor{c3}{HTML}{c40f40}

\definecolor{Gray}{gray}{0.93}
\newtheorem{theorem}{Theorem}[section]
\newtheorem{proposition}[theorem]{Proposition}

\theoremstyle{definition}

\theoremstyle{remark}

\usepackage{arydshln}

\usepackage[textsize=tiny]{todonotes}

\icmltitlerunning{Training Class-Imbalanced Diffusion Model Via Overlap Optimization}
\begin{document}

\twocolumn[
\icmltitle{Training Class-Imbalanced Diffusion Model Via Overlap Optimization}



\icmlsetsymbol{equal}{*}

\begin{icmlauthorlist}
\icmlauthor{Divin Yan}{fudan,ucm}
\icmlauthor{Lu Qi}{ucm}
\icmlauthor{Vincent Tao Hu}{lmu}
\icmlauthor{Ming-Hsuan Yang}{ucm,google}
\icmlauthor{Meng Tang}{ucm}
\end{icmlauthorlist}

\icmlaffiliation{fudan}{Fudan University}
\icmlaffiliation{ucm}{University of California Merced}
\icmlaffiliation{google}{Google Research}
\icmlaffiliation{lmu}{CompVis Group, LMU Munich}

\icmlcorrespondingauthor{Lu Qi}{qqlu1992@gmail.com}

\icmlkeywords{Machine Learning, ICML}

\vskip 0.3in
]



\printAffiliationsWithoutNotice{This work is done when Divin Yan is a visiting student at Vision and Learning Lab, University of California, Merced.}

\begin{abstract}

Diffusion models have made significant advances recently in high-quality image synthesis and related tasks. 
However, diffusion models trained on real-world datasets, which often follow long-tailed distributions, yield inferior fidelity for tail classes. 
Deep generative models, including diffusion models, are biased towards classes with abundant training images. 
To address the observed appearance overlap between synthesized images of rare classes and tail classes, we propose a method based on contrastive learning to minimize the overlap between distributions of synthetic images for different classes. 
We show variants of our probabilistic contrastive learning method can be applied to any class conditional diffusion model.
%
Extensive experimental results demonstrate that the proposed method can effectively handle imbalanced data for diffusion-based generation and classification models. Our code and datasets will be publicly available \href{https://github.com/yanliang3612/DiffROP}{here}.


\end{abstract}

\section{Introduction}







Diffusion models~\cite{sohl2015deep,ho2020denoising,song2021scorebased_sde} have facilitated generative models to synthesize high-quality image~\cite{rombach2022high}, 3D objects~\cite{poole2022dreamfusion}, and videos~\cite{yang2022diffusion}, to name a few. 
%
These diffusion models can be integrated with prompts such as text, class, semantic map, and sketch~\cite{zhang2023adding}, 
as conditions or guidance to render content effectively~\cite{azizi2023synthetic}.

Training diffusion models for image synthesis requires a large amount of data, which often follows a long-tailed distribution in the real world. 
Similar to discriminative models~\cite{jiang2021self,liu2022open} and other generative models such as Generative Adversarial Network (GAN)~\cite{tan2020fairgen}, diffusion models also perform worse for tail classes compared to head classes~\cite{qin2023class}.
%
Synthesized images are usually of low fidelity for rare classes. 
Data-driven generative models are shown to be biased to generate stereotypes~\cite{generativeaibias} which occur frequently in the training datasets. 
%
In this paper, we aim to improve the fairness and data efficiency of the class conditional diffusion model for image synthesis given long-tailed training data. 
Our goal is to improve image quality for tail classes while maintaining quality for head classes.

\begin{figure}[t]
\centering

\includegraphics[width=1.0\columnwidth]{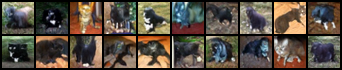} \\
\small{(a) Diffusion model generates \textit{cat}-like \textit{dog} images when trained on imbalanced dataset of 5000 \textit{cat} images and 50 \textit{dog} images.}
\includegraphics[width=0.8\columnwidth,trim={28cm 10cm 0 5cm},clip]{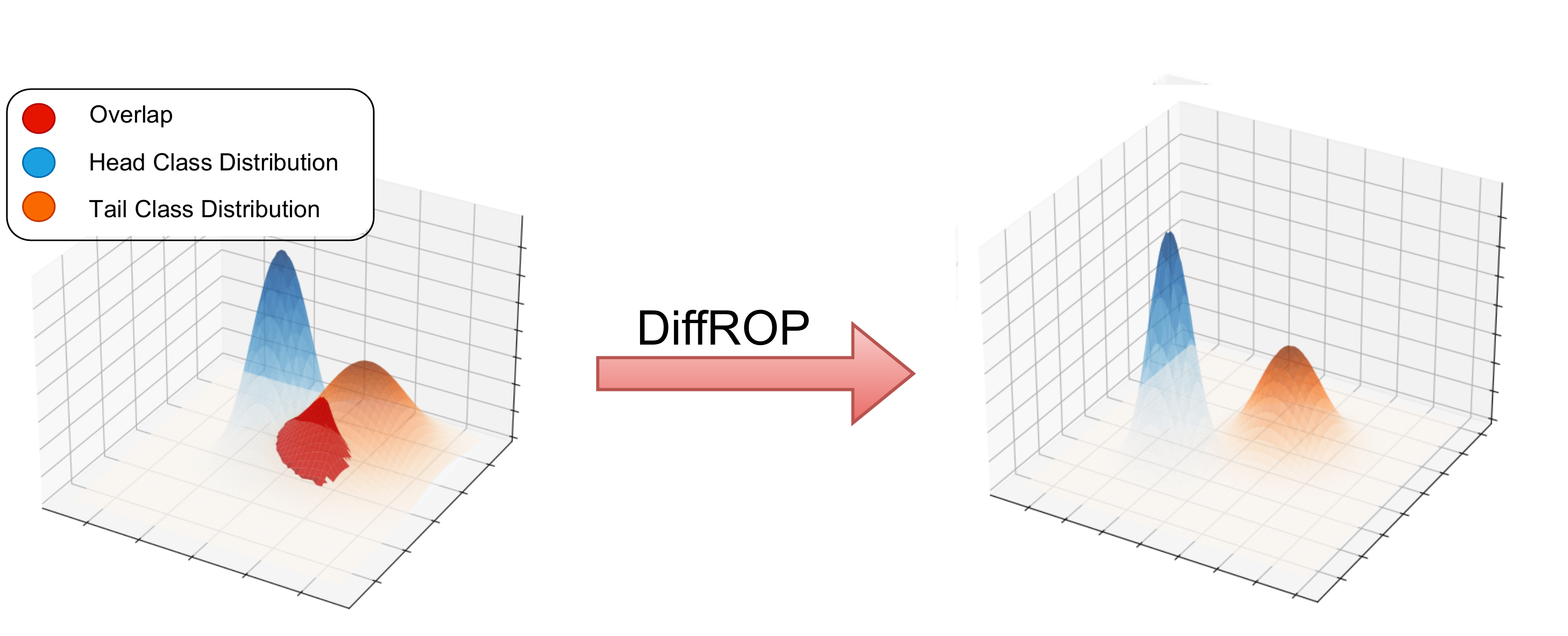} \\
(b) Our approach (DiffROP) minimizes overlap between the distributions of tail class images and rail class images.\\
\vspace{10pt}
\includegraphics[width=1.0\columnwidth,trim={0 0 0 0},clip]{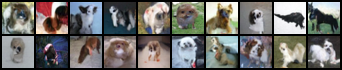} \\
(c) Generated rare class (\textit{dog}) images using our method.
\caption{Motivation of our method. 
}
\label{fig:apperanceoverlap}
\end{figure}

We observe that generated rare class images from a state-of-the-art diffusion model \cite{ho2020denoising,rombach2022high} tend to look like head class images (Fig. \ref{fig:apperanceoverlap}) for a model trained on two highly imbalanced classes. 
In other words, there is an overlap between the probabilistic distributions of generated rare class images and head class images. 
In this work, we propose to penalize such overlap between conditional image distributions via a Probabilistic Contrastive Learning (PCL) loss.
Specifically, we penalize KL divergence between conditional image distributions for different classes.
We show that such a KL divergence for distributions can be simply implemented by using estimated noises for two images.


Our method is highly related to contrastive learning which is a paradigm for representation learning by contrasting positive and negative instances. It has been shown effective for self-supervised learning~\cite{chen2020simple_simclr}, supervised learning~\cite{khosla2020supervised}, and long-tailed recognition~\cite{jiang2021self,liu2022open,li2022targeted}. 
Most relevant to this work is contrastive learning for long-tailed recognition. 
However, we note that this is the first work to introduce contrastive learning to diffusion models for long-tailed image synthesis. 
Unlike existing methods that learn per-instance embeddings through contrastive learning, we formulate a probabilistic contrastive learning framework to minimize 
Kullback–Leibler (KL) divergence of conditional image distributions for different classes.

The proposed method based on probabilistic contrastive learning is easy to implement and can be used for training any class conditional diffusion models. We effectively minimize PCL loss defined for image distributions by contrasting denoised images during training.
%
It does not require a tailored batch sampling scheme for positive and negative instances as required in other contrastive learning methods~\cite{khosla2020supervised,zhu2022balanced,chen2020simple_simclr}.
The PCL function is modular and can be combined with other components such as classifier free guidance~\cite{ho2021classifier} and class balancing diffusion model~\cite{qin2023class} with minimum computational overhead. 
We also explore several variants of probabilistic contrastive learning losses via visualization and thorough ablation study.

The main contributions of this work are:
\begin{itemize}
    \item We propose a method to handle imbalanced data for class conditional diffusion models via probabilistic contrastive learning. 
    The proposed loss function penalizes KL divergence between conditional image distributions for different classes. 
    %
    \item The proposed loss function is simple and efficient to implement using predicted noises for two images of different classes in a batch. 
    It can be easily incorporated as a regularization term with any diffusion model.
    Although this work is focused on image synthesis, the general loss can be used for any class conditional diffusion model.
    %
    %
    \item The proposed method significantly facilitates improving image fidelity for diffusion models trained on imbalanced datasets, particularly for tail classes.
    %
    Moreover, we demonstrate the proposed model can be effectively applied to generative data augmentation for downstream tasks including image classification.   
\end{itemize}

\section{Related Work}
\paragraph{Class-Imbalanced Representation Learning} 
Most real-world data is naturally imbalanced, presenting a significant challenge in training fair models that are not biased toward the majority classes. 
To address this problem, various approaches have been commonly utilized. 
Ensemble learning \cite{freund1997decision, liu2008exploratory, zhou2020bbn, wang2020long, liu2020mesa, cai2021ace} combines the outputs of multiple weak classifiers. 
Data re-sampling methods \cite{chawla2002smote, han2005borderline, smith2014instance, saez2015smote, kang2019decoupling, wang2021rsg} adjust the label distribution in the training set by synthesizing or duplicating samples from the minority class. 
Another approach tackles the imbalance issue by modifying the loss function, assigning larger weights to minority classes, or adjusting the margins between different classes \cite{zhou2005training, tang2008svms, cao2019learning, tang2020long, xu2020class, ren2020balanced, wang2021adaptive}. 
Post-hoc correction methods compensate for the imbalanced classes during the inference step, after completing the model training \cite{kang2019decoupling, tian2020posterior, menon2020long, hong2021disentangling}. 
Although these techniques have been extensively applied to discriminative tasks, extending them to generative models poses non-trivial challenges. 

\paragraph{Class-Imbalanced Generative Models} 
Generative models necessitate an extensive corpus of training data to handle the tasks well.
However, data in real-world scenarios often adhere to a long-tailed distribution.
Consequently, an increasing body of research~\cite{ roth2017stabilizing, zhao2020differentiable, karras2020training, tseng2021regularizing, mirza2021deep, ko2022self,qin2023class} converges to the exploration of methodologies to cultivate a robust generative model predicated on long-tailed or limited data distributions. 
A significant portion of these studies~\cite{roth2017stabilizing, zhao2020differentiable, karras2020training, tseng2021regularizing, mirza2021deep, ko2022self} primarily emphasizes data augmentation strategies, aiming to enhance the performance of GANs~\cite{goodfellow2020generative} or VAEs~\cite{kingma2022autoencoding}. CBDM~\cite{qin2023class} represents an inaugural inquiry into the performance of DDPM~\cite{ho2020denoising} within the context of long-tailed data scenarios. 
It introduces a data augmentation technique akin to label noise perturbation to augment model performance. 
Nevertheless, the enhancement in performance for minority classes remains notably insubstantial, and the model is prone to issues of gradient vanishing. 

\paragraph{Supervised Contrastive Learning for Class-Imbalance Problems}
Several studies~\cite{jiang2021self,li2022targeted} leverage contrastive learning to address long-tail recognition problems. Specifically,~\citet{jiang2021self} developed the Self-Damaging Contrastive Learning (SDCLR) framework, which ingeniously mitigates the imbalance inherent in long-tail distributions of unlabeled data by introducing a self-competitor model, thereby enhancing accuracy and achieving a more balanced performance across diverse scenarios. \citet{li2022targeted} proposed Targeted Supervised Contrastive Learning (TSC), an innovative approach that significantly enhances class uniformity and refines class boundaries within long-tailed recognition tasks, consequently setting new benchmarks in performance across several datasets. However, the aforementioned models predominantly address the long-tail issue in discriminative tasks. To the best of our knowledge, we are the first to adopt contrastive learning for the class-imbalanced diffusion model. It is noteworthy that our method operates at the level of probability distributions, rather than per-image embedding.

\section{Preliminaries}

\subsection{Unconditional Diffusion model}
A diffusion model leverages a pre-defined forward process in training, where a clean image distribution $q(x_0)$ can be corrupted to a noisy distribution $q(x_t|x_0)$ at a specified timestep $t$. 
Given a pre-defined variance schedule $\{\beta_t\}_{1:T}$, the noisy distribution at any intermediate timestep is 
\begin{equation}
        q(x_t | x_0) = \mathcal{N}(x_t;\sqrt{\bar \alpha_t} x_0, (1-\bar \alpha_t)\mI);~~~ \bar{\alpha}_t\!\!=\!\!\prod_{i=1}^t (1-\beta_i).
\end{equation}
To reverse such forward process, a generative model parameterized by $\theta$ learns to estimate the analytical true posterior in order to recover $x_{t-1}$ from $x_t$ as follows:
$$\min_\theta \KL[q(x_{t-1}|x_t, x_0)||p_{\theta}(x_{t-1}|x_t)];~~\forall t \in \{1, ..., T\},$$
and such an objective can be reduced to a simple denoising estimation loss~\cite{ho2020denoising}:
\begin{equation}\label{eq:mse}
\resizebox{\columnwidth}{!}{$\gL_\text{DDPM-UC} \!=\! \E_{t, x_0 \sim q(x_0), \rvepsilon \sim \gN(\vzero, \mI)} \left[ \|\rvepsilon - \rvepsilon_{\theta}(\sqrt{\bar{\alpha}_t}x_0+\sqrt{1-\bar{\alpha_t}}\rvepsilon, t)\|^2 \right]$}. 
\end{equation}

\subsection{Conditional Diffusion Model}

For the case where class label $\mathbf{c}$ is available for image $x_0$, the training objective is:
\begin{equation}
    \begin{aligned}
        \mathcal{L} &= \mathbb{E}_{q}\Biggl\{-\text{log} p_{\theta}(x_{0}|x_{1},\mathbf{c}) + D_{KL}\left[q(x_{T}|x_{0},\mathbf{c})||p(x_{T})\right] \\
        &+ \sum_{t\geq1}^{T}D_{KL}\left[q(x_{t-1}|x_{t},x_{0},\mathbf{c})||p_{\theta}(x_{t-1}|x_{t},\mathbf{c})\right]
       \Biggr\}.
    \end{aligned}
\end{equation}
Similar to \citet{ho2020denoising},  the training objective can be simplified to
\begin{equation}
    \sum_{t\geq1}^{T}D_{KL}\left[q(x_{t-1}|x_{t},x_{0},\mathbf{c})||p_{\theta}(x_{t-1}|x_{t},\mathbf{c})\right]. 
\label{eq:kl_loss}
\end{equation}

The simple denoising estimation loss in \Eqref{eq:mse} becomes:
\begin{equation}\label{eq:mse_conditioned}
\resizebox{\columnwidth}{!}{$\gL_\text{DDPM-C} \!=\! \E_{t, x_0 \sim q(x_0), \rvepsilon \sim \gN(\vzero, \mI)} \left[ \|\rvepsilon - \rvepsilon_{\theta}(\sqrt{\bar{\alpha}_t}x_0+\sqrt{1-\bar{\alpha_t}}\rvepsilon, t, \mathbf{c})\|^2 \right]$}.
\end{equation}

\textbf{Classifier-free guidance} The model is trained to estimate the noise in both conditional cases $\rvepsilon_{\theta}(x_t, c, t)$ with data-label pairs $(x_0,\mathbf{c})$ and unconditional case $\rvepsilon_{\theta}(x_t, t)$.
In the sampling process, the label-guided model estimates the noise with a linear interpolation $\hat \rvepsilon = (1+\omega)\rvepsilon_{\theta}(x_t, c, t) - \omega \rvepsilon_{\theta}(x_t, t)$ to recover $x_{t-1}$, which is often referred as Classifier-Free Guidance (CFG)~\cite{ho2021classifier}. 

\subsection{Class-imbalanced Diffusion model}
The number of training images for each class is significantly different for the class-imbalanced diffusion model.
Let $\mathcal{C}$ be the set of classes and $w_c$ be the portion of training images for class $\mathbf{c}$, i.e., $\sum_c{w_c}=1$, the original KL divergence can be seen as a weighted sum of KL divergence for each class.

\begin{proposition}[\textbf{Weight-Biased Decomposition of DDPM Loss Function}]
The original training objective in \Cref{eq:kl_loss} for DDPM can be rewritten as
\begin{equation}
    \begin{aligned}
        &\mathbb{E}_{q}\Biggl\{\sum_{t\geq1}^{T}D_{KL}\left[q(x_{t-1}|x_{t},x_{0},\mathbf{c})||p_{\theta}(x_{t-1}|x_{t},\mathbf{c})\right]\Biggr\}\\
        &= \sum_{c \in \mathcal{C}} w_c \mathbb{E}_{q} \Biggl\{\sum_{t\geq1}^{T} D_{KL}\left[q(x_{t-1}|x_{t},x_{0},\mathbf{c})||p_{\theta}(x_{t-1}|x_{t},\mathbf{c})\right]\Biggr\}.
    \end{aligned}
\label{eq:class_weighted_ddpm_loss}
\end{equation}
\label{prop:weighted_kl}
\end{proposition}
The proof is presented in Appendix \ref{proof_by_class}. 
From \Eqref{eq:class_weighted_ddpm_loss}, it is clear that DDPM is biased toward head classes with large $w_c$ leading to observed overlap between head class images and tail class images. 
A naive solution is to weigh the loss for each training image $x_0$ using a weight inversely proportional to $w_c$. 
%
However, such a weighted loss does not work in practice as shown in Table \ref{tab:weighted_loss}.

\section{Method}
In this section, we introduce our \textbf{DiffROP} (\underline{\textnormal{Diff}}usion framework with $\underline{\textnormal{R}}$egularize $\underline{\textnormal{O}}$verla$\underline{\textnormal{P}}$) framework which is motivated by minimizing the overlap between class-conditional distributions.

In Section \ref{sec:pclloss}, we introduce a loss based on probabilistic contrastive learning (PCL) to penalize the KL divergence between distributions for different classes.
We show how this loss can be effectively minimized via estimated noises for two classes.
In Section \ref{sec:overall_framework}, we integrate this PCL loss into the original diffusion model loss and show our overall framework, providing an effective approach to improve the performance of class-imbalanced diffusion models.

%
%

\subsection{Probabilistic Contrastive Learning Loss}
\label{sec:pclloss}
%
With the motivation of distinctly demarcating the distribution boundaries of different data categories, we propose a loss to penalize the KL divergence of probability distributions for different classes. Our loss contrasts sampled data from two classes and is based on conditional distributions rather than latent embeddings. Hence, our loss is termed a probabilistic contrastive learning loss.

%
%
Specifically, for two different classes, ${c}^{i},\ \text{and} \ {c}^{j}$, we randomly sample two images $x_{0}^i$ and ${x}_{0}^j$, one from class ${c}^i$ and the other from class ${c}^{j}$. 
We penalize KL divergence between two conditional distributions $p_{\theta}(x_{t-1}|x_{t}^i,\boldsymbol{c}^i)$ and  $p_{\theta}({x}_{t-1}|{x}_{t}^j,\boldsymbol{c}^j)$, 
\begin{equation}
    \begin{aligned}
        \mathcal{L}_{PCL}^{ij,t} &=D_{KL}\left[p_{\theta}(x_{t-1}|x_{t}^i,\boldsymbol{c}^i)||p_{\theta}({x}_{t-1}|{x}_{t}^j,\boldsymbol{c}^j)\right]. 
    \end{aligned}
\end{equation}

By penalizing distributions overlap, we can enjoy higher fidelity for tail classes due to less appearance overlap with images of head classes. Next, we show how the KL divergence loss $\mathcal{L}_{PCL}^{ij,t}$ can be reformulated using a noise estimator for two classes $\mathbf{c}^i$ and $\mathbf{c}^j$. Suppose $p_{\theta}(x_{t-1}|x_{t},\boldsymbol{c}) = \mathcal{N}(x_{t-1};\mu_{\theta}(x_{t},t,\boldsymbol{c}),\sigma_{t}^{2}\mathbf{I})$, we can write:
\begin{equation}
    \begin{aligned}
        \mathcal{L}_{PCL}^{ij,t} &= \mathbb{E}_{q}\left[\frac{1}{2\sigma_{t}^{2}}\lVert\ \mu_{\theta}(x_{t}^i,t,\boldsymbol{c}^{i})-\mu_{\theta}({x}_{t}^j,t,\boldsymbol{c}^{j})\rVert^{2}\right] + C,
    \end{aligned}
    \label{111}
\end{equation}
where $C$ is a constant and
\begin{equation}
\begin{aligned}
    \mu_{\theta}\left(x_{t}^i, t, ,\boldsymbol{c}^{i}\right)&=\frac{1}{\sqrt{\alpha_{t}}}x_{t}^i\\
    &-\frac{1-\alpha_{t}}{\sqrt{\alpha_{t}}\sqrt{1-\bar{\alpha}_{t}}} \epsilon_{\theta}\left(x_{t}^i, t, \boldsymbol{c}^{i}\right),
\end{aligned}
\end{equation}
\begin{equation}
\begin{aligned}
    \mu_{\theta}\left({x}_{t}^j, t, ,\boldsymbol{c}^{j}\right)&=\frac{1}{\sqrt{\alpha_{t}}}
    {x}_{t}^j\\
    &-\frac{1-\alpha_{t}}{\sqrt{\alpha_{t}}\sqrt{1-\bar{\alpha}_{t}}} \epsilon_{\theta}\left({x}_{t}^j, t, \boldsymbol{c}^{j}\right). 
\end{aligned}
\end{equation}
%
Thus,  $\mathcal{L}_{PCL}^{ij,t}$ can be simplified to: 
\begin{equation}
\left\lVert (x_t^i - {x}_t^j) + \frac{1-\alpha_{t}}{\sqrt{1-\bar{\alpha}_{t}}} (\epsilon_{\theta}\left({x}_{t}^j, t, \boldsymbol{c}^{j}\right) - \epsilon_{\theta}\left(x_{t}^i, t, \boldsymbol{c}^{i}\right)) \right\rVert^{2}. 
\end{equation}
Based on $x_{t}(x_{0},\epsilon) = \sqrt{\overline{\alpha}_{t}}x_{0} + \sqrt{1-\overline{\alpha}_{t}}\epsilon, \epsilon \sim \mathcal{N}(0,\mathbf{I})$, with the parameterization trick , $\mathcal{L}_{PCL}^{ij,t}$ simplifies to:
\begin{equation}
    \begin{aligned}
        & \mathbb{E}_{x_{0},\epsilon}\Biggl[\Bigl\lVert (x_t^i - {x}_t^j) + \frac{1-\alpha_{t}}{\sqrt{1-\bar{\alpha}_{t}}} \Bigl(\epsilon_{\theta}\bigl(\sqrt{\overline{\alpha}_{t}} {x}_{0}^j+\sqrt{1-\overline{\alpha}_{t}}\epsilon,t,\boldsymbol{c}^{j}\bigr)\\
        &-\epsilon_{\theta}\bigl(\sqrt{\overline{\alpha}_{t}} x_{0}^i+\sqrt{1-\overline{\alpha}_{t}}\epsilon,t,\boldsymbol{c}^{i}\bigr)\Bigr)\Bigr\rVert^{2}\Biggr]. 
    \end{aligned}
\end{equation}
In other words, our proposed PCL loss $\mathcal{L}_{PCL}^{ij,t}$ can be effectively reformulated using denoised images $\mu_{\theta}(x_{t},t,\boldsymbol{c})$, which add little computational overhead for training.


\subsection{Overall Loss and Framework}
\label{sec:overall_framework}

%
A naive way of integrating our probabilistic contrastive learning loss is to sum the original diffusion model loss and PCL loss. For a dataset with classes $\mathcal{\boldsymbol{C}}$, the overall training objective becomes:
\begin{equation}
    \begin{aligned}
        \mathcal{L}_{\text{overall}} &= \sum_{t \geq1}\underbrace{(D_{KL}\left[q(x_{t-1}|x_{t},x_{0})||p_{\theta}(x_{t-1}|x_{t},\boldsymbol{c})\right])}_{\text{DDPM loss}} \\
        &- \tau \cdot \mathbb{E}_{{\mathbf{c}^i\in \mathcal{C},\mathbf{c}^j\in \mathcal{C},\mathbf{c}^i\neq \mathbf{c}^j}} \\
        &\Biggl[\sum_{t \geq 1}\underbrace{D_{KL}\left[p_{\theta}(x_{t-1}|x_{t}^i,\boldsymbol{c}^{i})|| 
        p_{\theta}({x}_{t-1}|{x}_{t}^j,\boldsymbol{c}^j)\right]}_{\text{Probabilistic Contrastive Learning Loss} \ \mathcal{L}_{PCL}^{ij,t}}\Biggr],\\
    \end{aligned}
\end{equation}
where $\tau$ is a weight hyperparameter. 

The corresponding loss with the noise estimator is
\begin{equation}
\begin{aligned}
 & \mathcal{L}_{\text{overall}}^{\text{simple}} = \E_{t, x_0 \sim q(x_0), \rvepsilon } \left[ \|\rvepsilon - \rvepsilon_{\theta}(\sqrt{\bar{\alpha}}x_0+\sqrt{1-\bar{\alpha_t}}\rvepsilon, t, c)\|^2 \right] \\
 & -\tau \cdot E_{t, x_0^i \sim q(x_0), x_0^j \sim q(x_0), \rvepsilon } \left[\lVert\ \mu_{\theta}(x_{t}^i,t,\boldsymbol{c}^{i})-\mu_{\theta}({x}_{t}^j,t,\boldsymbol{c}^{j})\rVert^{2}\right]. \\
\end{aligned}
\end{equation}
Our overall loss above is simple to implement. However, naive implementation with proposed PCL loss leads to worse performance compared to the original diffusion model as shown in Section \ref{sec:ablation}. Here, we introduce two ways of improving our losses that are critical for our method to be effective.

\textbf{Time-dependent $\tau$.} In the forward diffusion process, images are corrupted toward a Gaussian distribution. We anticipate more distributions overlap for different classes with increasing timestamp $t$. Hence, we propose decreasing loss weight $\tau$, e.g., temperature-controlled exponential decay w.r.t. timestamp.

\textbf{Hinge-like loss.} To encourage a margin between probability distributions for two classes while keeping $p_\theta=q$ as the global optima for the overall objective, we propose a Hinge loss for two conditional distributions.
\begin{equation}
        \max(0, D_t - D_{KL}\left[p_{\theta}(x_{t-1}|x_{t}^i,\boldsymbol{c}^i)||p_{\theta}({x}_{t-1}|{x}_{t}^j,\boldsymbol{c}^j)\right]),
\end{equation}
where $D_t$ is the margin for divergence for timestamp $t$. It is obvious to see that the Hinge loss for KL divergence translates to a Hinge loss for denoised images. We propose multiple variants of losses in our study, which are visualized in Section \ref{sec:loss_visualization} for toy data:
\begin{itemize}
  \item Negative $L_2$ distance: 
  \item[] $-\lVert\ \mu_{\theta}(x_{t}^i,t,\boldsymbol{c}^{i})-\mu_{\theta}({x}_{t}^j,t,\boldsymbol{c}^{j})\rVert^{2}$
    \item Max margin Hinge loss with margin $\mu_t$: 
    \item[]$\max(0, \mu_t - \lVert\ \mu_{\theta}(x_{t}^i,t,\boldsymbol{c}^{i})-\mu_{\theta}({x}_{t}^j,t,\boldsymbol{c}^{j})\rVert^{2}$ .
    \item Hinge-like loss (Reciprocal form): 
    \item[]$\frac{1}{1+\lVert\ \mu_{\theta}(x_{t}^i,t,\boldsymbol{c}^{i})-\mu_{\theta}({x}_{t}^j,t,\boldsymbol{c}^{j})\rVert^{2}}$.
    \item Hinge-like loss (Exponential form):
    \item[] $e^{-\lVert\ \mu_{\theta}(x_{t}^i,t,\boldsymbol{c}^{i})-\mu_{\theta}({x}_{t}^j,t,\boldsymbol{c}^{j})\rVert^{2}}$.
\end{itemize}

Our framework doesn't require a special positive/negative pair sampling scheme. We adopt randomly sampled batches from the original diffusion model. Algorithm \ref{alg:training} outlines our framework.

\begin{algorithm}[t!]
\caption{Training algorithm of DiffROP.}\label{xt1_algo}
\begin{algorithmic}[1]
    \FOR {each batch}
        \FOR {each image-class pair $(x_0^{i}, c^{i})$ in this batch}
            \STATE Sample $\rvepsilon^i \sim \mathcal{N}(\vzero, \mI)$, $t\sim \mathcal{U}(\{0, 1, ..., T\})$
            \STATE $\rvx^{i}_{t} = \sqrt{\bar{\alpha}_{t}}\rvx^{i}_0+ \sqrt{1-\bar{\alpha}_{t}}\rvepsilon^i$
            \STATE \# Original diffusion model loss
            \STATE Compute $ \gL_\text{DM} = \| \rvepsilon^{i} - \rvepsilon_{\theta}(x^{i}_{t}, t, c^{i})\|^2$
            \STATE Set $\mathcal{L}_{PCL} = 0$
            \STATE \# Probabilistic Contrastive Learning loss
            \FOR {$(\rvx_0^{j}, c^{j})$ in this batch with $c^i\neq c^j$}
                \STATE \# Compute squared distance
                \STATE $d_{ij}=\lVert\ \mu_{\theta}(x_{t}^i,t,\boldsymbol{c}^{i})-\mu_{\theta}({x}_{t}^j,t,\boldsymbol{c}^{j})\rVert^{2}$
                \STATE \# Compute hinge-like loss
                \STATE $\mathcal{L}_{PCL} = \mathcal{L}_{PCL} + \tau(t) \cdot \frac{1}{1+d_{ij}}$
            \ENDFOR

            \STATE Update with $\gL_\text{DiffROP} = \gL_\text{DM} + \gL_\text{PCL}$
        \ENDFOR
    \ENDFOR
\end{algorithmic}
\label{alg:training}
\end{algorithm}

%

\section{Experiments}

Firstly, we visualize our loss using toy data for a basic understanding. Secondly, we show our results on major benchmark datasets, followed by in-depth ablation studies for different variants of our methods and different hyperprameter settings.

\subsection{Loss visualization on toy data}
\label{sec:loss_visualization}

\begin{figure}[t!]
\centering

\includegraphics[width=0.99\columnwidth]{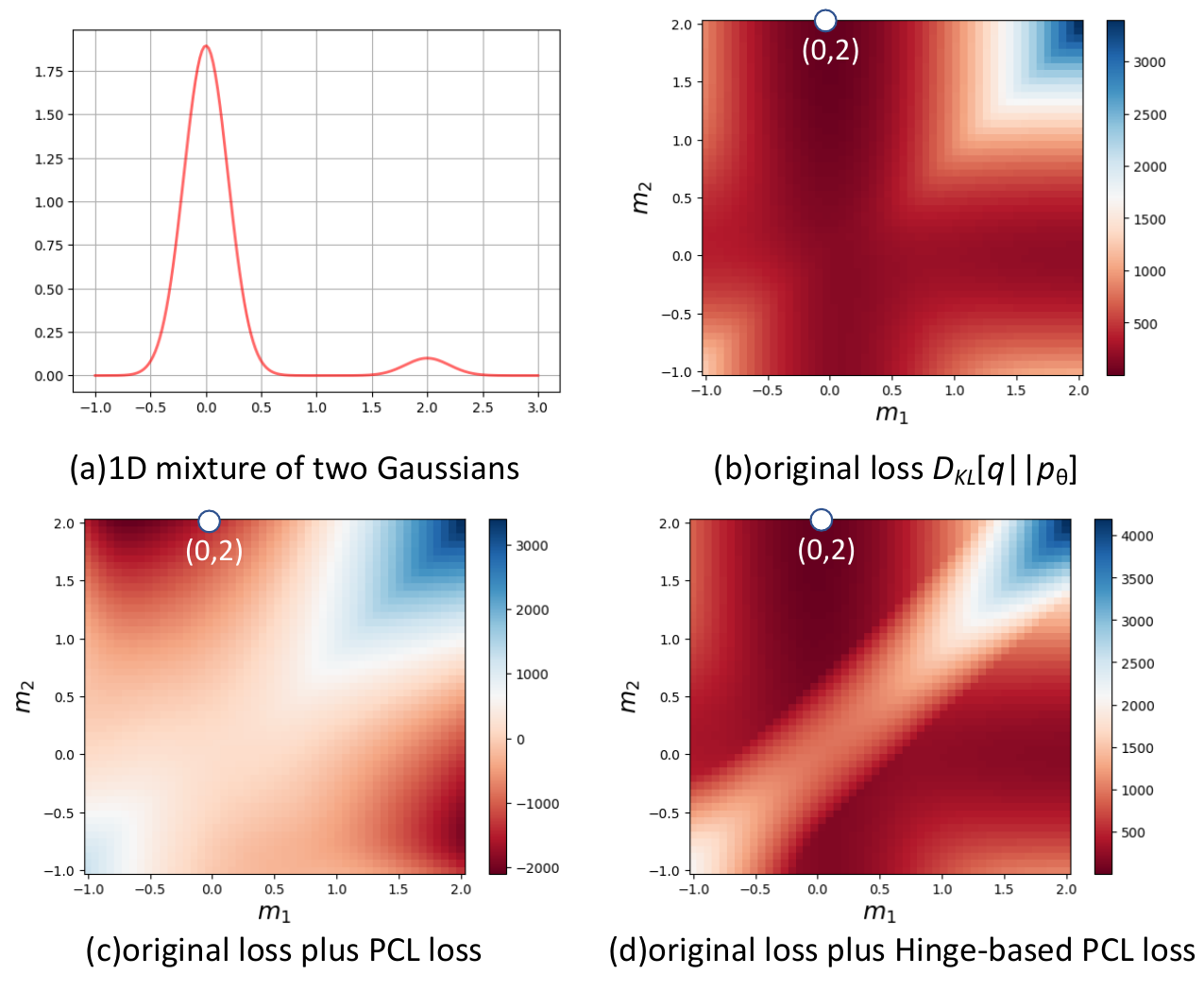} \\
\caption{Loss visualization w.r.t. the two estimated means of Guassians $m_1$ and $m_2$. With our Hinge-based PCL loss, solutions with large distribution overlap (when $m_1$ is close to $m_2$) will be penalized, while the global optima $(0,2)$ is preserved.}
\label{fig:loss_visualization}
\end{figure}

To understand the effect of the proposed loss, we visualize losses in Figure \ref{fig:loss_visualization} for 1D toy data. We assume an underlying data distribution of a mixture of two Gaussians

\begin{equation}
    q(x) = \pi_1 \mathcal{N}(x;m_1^\ast,\sigma_1^2)  + \pi_2 \mathcal{N}(x;m_2^\ast,\sigma_2^2),
\end{equation}

with imbalanced weights $\pi_1=0.95,\pi_2=0.05$ and $m_1^\ast=0,m_2^\ast=2$ (Figure \ref{fig:loss_visualization} (a)). Assuming known weights $\pi$ and variance $\sigma^2$, we visualize losses w.r.t. means $m_1,m_2$ to be estimated. Naively adding PCL loss $-D_{KL}\left[\mathcal{N}(x;m_1,\sigma^2) || \mathcal{N}(x;m_2,\sigma^2)\right]$ will change the global optima (Figure \ref{fig:loss_visualization} (c)). Using Hinge-based PCL loss minimizes distribution overlap while preserving global optima of $m_1=0,m_2=2$ (Figure \ref{fig:loss_visualization} (d)), see discussion of Hinge-like losses in Section \ref{sec:overall_framework}.

\begin{table*}[htb!]
  \centering
  \caption{
  Performance of evaluated methods on benchmarks.  Highlighted are the top \textbf{\textcolor{c1}{first}} and \textbf{\textcolor{c2}{second}}.}
  \label{tab:main}
  \begin{tabular}{lllllll}
    \toprule[1.5pt]
    Dataset & Model & \textbf{FID}$\downarrow$ &  $\bm{F_{8}}$ $\uparrow$&\textbf{Recall}$\uparrow$ & \textbf{IS}$\uparrow$  & $\bm{F_{1/8}}$ $\uparrow$ \\
    \midrule
    
    CIFAR100LT & DDPM~\cite{ho2020denoising} & \textbf{\textcolor{c2}{7.38}}	& 0.85  & \textbf{\textcolor{c2}{0.52}}   &\textcolor{c2}{\textbf{13.11}}&  \textbf{\textcolor{c2}{0.88}} \\
    &  ~+DiffAug~\cite{zhao2020differentiable}  & 9.19  & \textbf{\textcolor{c2}{0.88}}  &  0.47   & 11.56 & 0.86 \\    
    & ~+ RS~\cite{mahajan2018exploring} & 10.50 & 0.65    & 0.49   & 12.60 & 0.83 \\
    & ~+SQRT- RS~\cite{mahajan2018exploring} & 9.72  & 0.66  & 0.47    & \textcolor{c1}{\textbf{13.47}} & 0.83 \\
        & ~{+DiffROP} & \textbf{\textcolor{c1}{6.84}}  &\textbf{\textcolor{c1}{0.89}}   & \textbf{\textcolor{c1}{0.56}}    &12.88  & \textbf{\textcolor{c1}{0.90}} \\
    \cline{2-7}
    & CBDM~\cite{qin2023class}  & 6.26 & 0.91 &   0.57  &   13.24  & 0.89  \\
    & ~{+DiffROP}  &\textbf{\textcolor{c1}{5.99}}  &0.89  &\textcolor{c1}{\textbf{0.57}}     &\textcolor{c1}{\textbf{13.24}}    &\textcolor{c1}{\textbf{0.91}}  \\
    \midrule
    CIFAR10LT & DDPM~\cite{ho2020denoising} & 5.76& 0.97 & 0.57    &9.17 & 0.95 \\
        & ~{+DiffROP} &\textbf{\textcolor{c1}{5.34}}   &0.95   &\textbf{\textcolor{c1}{0.58}}     &\textbf{\textcolor{c1}{9.18}}  &\textbf{\textcolor{c1}{0.97}} \\
    \cline{2-7}
    & CBDM~\cite{qin2023class} &5.46  & 0.97  &  0.59 &  9.28 &  0.95 \\
    & ~{+DiffROP} &\textbf{\textcolor{c1}{5.35}}  &0.95   &\textbf{\textcolor{c1}{0.60}}   &9.19   &\textbf{\textcolor{c1}{0.97}}   \\
    \bottomrule[1.5pt]
  \end{tabular}
  
\end{table*}
\subsection{Experimental Setup}

\paragraph{Datasets}
We initially selected a pair of prevalent datasets, CIFAR10/CIFAR100~\cite{cao2019learning}, commonly adopted in image synthesis studies and long-tailed recognition. Alongside, we also employed their skewed versions, namely CIFAR10LT~\cite{cao2019learning} and CIFAR100LT~\cite{cao2019learning}. The formulation of CIFAR10LT and CIFAR100LT is aligned with the methodology presented by \citet{cao2019learning, qin2023class}. Specifically, the dataset size diminishes exponentially based on the category index, guided by the imbalance factor $\textit{imb}=0.01$. 

\begin{table}[tb]
\setlength{\tabcolsep}{2.5pt}
\small
\begin{center}
\caption{\textbf{FID} Scores for each interval on Benchmarking results. Highlighted are the top \textbf{\textcolor{c1}{first}}.}
\label{diff_class}
\begin{tabular}{l|ccc|ccc}
\toprule[1.1pt]
Datasets & \multicolumn{3}{c|}{CIFAR10LT} & \multicolumn{3}{c}{CIFAR100LT} \\ 
\midrule
Shot & \textit{Many} & \textit{Med} & \textit{Few} & \textit{Many} & \textit{Med} & \textit{Few} \\ 
\midrule
\textsc{DDPM} & 28.82 & 37.01 & 46.51 & 23.30 & 27.90 & 30.43 \\
    {~+DiffROP} & \textcolor{c1}{\textbf{23.51}} & \textcolor{c1}{\textbf{33.63}} & \textcolor{c1}{\textbf{39.23}} & \textcolor{c1}{\textbf{21.73}} & \textcolor{c1}{\textbf{27.50}} & \textcolor{c1}{\textbf{29.13}} \\
\midrule
\textsc{CBDM} & 26.18 & 37.53 & 47.01 & 21.99 & 27.40 & 28.33 \\ 
        {~+DiffROP} & \textcolor{c1}{\textbf{21.77}} & \textcolor{c1}{\textbf{36.00}} & \textcolor{c1}{\textbf{36.65}} & \textcolor{c1}{\textbf{21.52}} & \textcolor{c1}{\textbf{26.81}} & \textcolor{c1}{\textbf{27.10}} \\ 
\bottomrule[1.1pt]
\end{tabular}

\end{center}
\vspace{-0.1cm}
\end{table}

\vspace{-12pt} 
\paragraph{Technical Specifications} 
Our approach adheres closely to the training protocols of reference models. For the DDPM model, we configure the diffusion schedule~\cite{ho2020denoising} with parameters $\beta_1={10}^{-4}$ and $\beta_T=0.02$, using $T=1{,}000$. The Adam optimizer~\cite{kingma2014adam} is employed for network optimization, utilizing a learning rate of $0.0002$ post a warmup of 5,000 steps.

\vspace{-12pt} 
\paragraph{Evaluation Criteria} 
The performance of methods is gauged using metrics that account for both the diversity and fidelity of generation. These metrics comprise the Fr\'echet Inception Distance (FID) \cite{heusel2017gans}, Inception Score (IS) \cite{salimans2016improved}, Recall \cite{kynkaanniemi2019improved}, and ${F}_{\beta}$ \cite{sajjadi2018assessing}. For Recall and ${F}_{\beta}$ calculations, we employ Inception-V3 features, adopting parameters $K=5$ for Recall and respective values of 1/8 and 8 for the ${F}_{\beta}$ threshold. Furthermore, the clustering count for ${F}_{\beta}$ is set at 20 times the class count to ascertain intra-class variations. Consequently, metrics such as Recall and ${F}_{8}$ lean towards assessing diversity, whereas IS and ${F}_{1/8}$ are more fidelity-centric. 
In our evaluations, the datasets mirror the class balance  during training, and metric evaluations are performed on a set of 50k synthesized images (or 10k for segmented studies). The evaluation also incorporates classifier-free guidance during sampling, adjusting the guidance strength $\omega$ across all models to maximize results. 

\vspace{-12pt} 
\paragraph{Baseline Methods} 
We compare our DiffROP method with multiple baseline approaches including re-sampling (RS)~\cite{mahajan2018exploring}, the more nuanced soft re-sampling variant (RS-SQRT) ~\cite{mahajan2018exploring}, and augmentation strategies like Differentiable Augmentation (DiffAug)~\cite{zhao2020differentiable}. Beyond these diffusion-centric baselines, we also compare to leading-edge generative models focused on long-tailed distributions, namely CBGAN \cite{rangwani2021class}, and the group spectral regularization mechanism for GANs \cite{goodfellow2014generative}. Specifically, RS is characterized by a homogenous class distribution while RS-SQRT~\cite{mahajan2018exploring} utilizes a probability derived from the square root of class frequencies. DiffAug~\cite{zhao2020differentiable} is uniformly applied across all training visuals. As per \cite{karras2022elucidating} extends beyond image application, encoding the augmentation pipeline conditionally via an auxiliary embedding layer in U-Net.

\subsection{Main Results}

\paragraph{Main Results On Imbalance Datasets.}
Table~\ref{tab:main} showcases the performance of the DiffROP model on CIFAR100LT and CIFAR10LT datasets. The DiffROP model, which incorporates an optimization for distribution overlap in diffusion models, has shown promising results. In CIFAR100LT, the DiffROP model has reduced the FID score from 7.38 to 5.99, indicating that it generates images closer to the real data, even though they are long-tail distributed. The Recall has increased to 0.57, meaning the model captures the data's diversity better in these special regimes. The Inception Score is high at 13.23, showing that the generated images are both clear and varied. The $F_{1/8}$ score is the highest at 0.91, suggesting a good balance of precision and recall. For CIFAR10LT, DiffROP also improves the FID score to 5.35 and maintains a high Recall of 0.60 and a good Inception Score of 9.19. The $F_{1/8}$ score remains very high at 0.97. Overall, the implementation of the DiffROP model has significantly boosted both the quality and the heterogeneity of the synthesized images, underscoring its efficacy as a strategy for augmenting the performance of diffusion models, particularly in scenarios characterized by dataset imbalance.

\subsection{Results Across Various Category Intervals}

\paragraph{Experimental Setup}
To better dissect the performance among various classes we preprocess the dataset, we initially divided it into three distinct categories: \textit{many}, \textit{med}, and \textit{few}. This division was carried out by equally distributing the dataset based on the size of each class, which was sorted in descending order. For the CIFAR10LT dataset, the top 3 classes were designated for the \textit{many} category, the subsequent 4 classes for the \textit{med} category, and the remaining 3 classes for the \textit{few} category. Similarly, for the CIFAR100LT dataset, the top 33 classes were allocated to the \textit{many} category, the next 34 classes to the \textit{med} category, and the last 33 classes to the \textit{few} category. Following the partitioning, we computed and stored the statistics and embeddings separately with respective to each category from the training data. Specifically for FID calculation, the mean and covariance were calculated using a pre-trained InceptionV3~\cite{szegedy2016rethinking_inceptionv3}, and embeddings from InceptionV3 were preserved for precision and recall computation. Throughout the training phase, we strictly adhered to the hyperparameters prescribed by the authors of CBDM. Subsequently, during the evaluation stage, we segregated the generated samples based on their respective class categories and conducted assessments independently for each category. All experiments were carried out utilizing 8 RTX 3090 GPUs.

\paragraph{Results Analysis} Table~\ref{diff_class} presents a detailed evaluation of the DiffROP framework, integrated with both DDPM and CBDM. This assessment spans various class categories, differentiated by the number of samples per class. In the context of the CIFAR10LT dataset, DiffROP significantly enhances performance across the \textit{many}, \textit{medium}, and \textit{few} sample categories. This improvement is particularly salient in the context of tail classes, highlighting DiffROP's versatility and effectiveness. Likewise, in the CIFAR100LT dataset, DiffROP achieves substantial improvements in FID scores across all categories. These results demonstrate DiffROP's capability to generate high-quality images consistently, irrespective of class quantity or dataset imbalances.

To further gain insights into the generation and performance dynamics of DiffROP, we conducted a meticulous comparison of FID scores, examining each case individually between DDPM and CBDM when coupled with DiffROP. The evaluation spanned a comprehensive assessment across three distinct class categories, meticulously divided based on their sizes. The ensuing results presented in Table \ref{diff_class}, revealed a notable trend in scenarios where DiffROP was integrated, consistent enhancements across all three categories were observed compared to instances without its incorporation, showcasing its robust effect. Particularly noteworthy were the outcomes in the tail classes, where DiffROP exhibited a pronounced superiority over both vanilla DDPM and CBDM, boasting a substantially wider margin in FID scores.

\begin{figure}[bt]
\centering

\includegraphics[width=1.0\columnwidth,trim={0 7.2cm 0 0},clip]{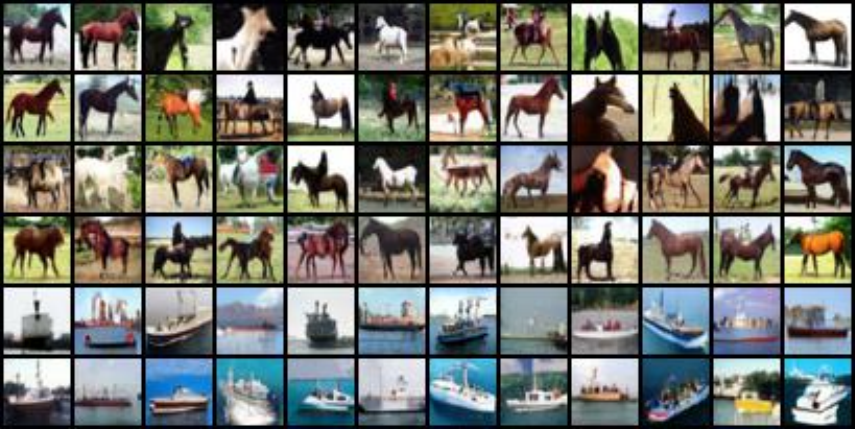} \\
\small{(a) DDPM} \\
\includegraphics[width=1.0\columnwidth,trim={0 7.2cm 0 0},clip]{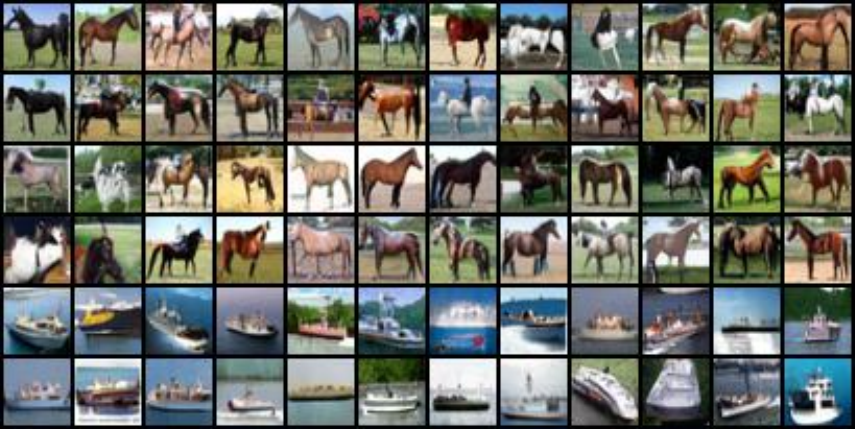} \\
(b) CBDM\\
\includegraphics[width=1.0\columnwidth,trim={0 7.2cm 0 0},clip]{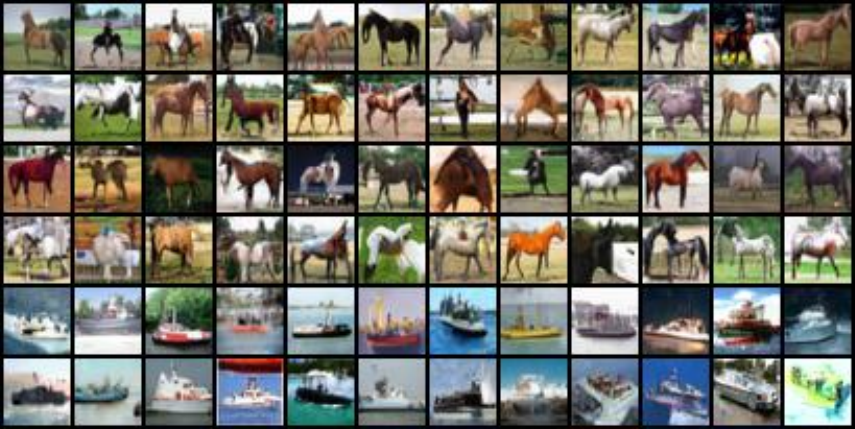} \\
(c) DiffROP
\caption{Qualitative results for tail classes. Our method is better at creating clear and realistic images for less common classes compared to other basic methods.}
\label{fig:tail_class_images}
\vspace{-0.4in}
\end{figure}

\vspace{-12pt}

\subsection{Data Augmentation for the Long-tail Classifiers}
In Table~\ref{tab:recog}, our investigation predominantly concentrates on the augmentation of classifier efficacy within the realm of long-tailed data distributions, an endeavor that is both arduous and paramount within the domain of machine learning. 
Our innovative approach combines advanced techniques with DDPM~\cite{ho2020denoising} and CBDM~\cite{qin2023class}, leading to significant improvements in performance metrics. 
This pronounced enhancement is indicative of our generative model's capability to synthesize data characterized by more delineated class demarcations. When this synthetically generated corpus is employed in downstream applications, such as classification tasks, it engenders a performance that is not only more resilient but also markedly efficacious. This underscores the pragmatic utility of our sophisticated approach amidst the intricacies of intricate data landscapes.

\begin{table}[t]
  \centering
  \caption{Recognition results($\%$) of different training data. 
  }
  \label{tab:recog}
  \begin{tabular}{l| lll}
    \toprule
    Training Data &Accuracy& Precison & Recall \\
    \midrule
    \textcolor{gray}{{CIFAR100}} &\textcolor{gray}{69.80}  & \textcolor{gray}{69.10}  & \textcolor{gray}{68.87}  \\ 
     \midrule
    CIFAR100LT & 37.74 & 42.12	& 37.54 \\
    ~+ DDPM gens (50k)  &44.08  & 48.22  &43.58  \\
    ~ {+DiffROP gens (50k)}  &\textbf{46.41}  & \textbf{48.22}  &\textbf{45.54} \\
    \midrule
    ~+ CBDM gens (50k)  &46.01  & 48.95   & 45.46 \\
    ~ {+DiffROP gens (50k)} &\textbf{47.22}   &\textbf{49.13}   &\textbf{46.76} \\
    \bottomrule
  \end{tabular}
\end{table}

\subsection{DiffROP and Reweighting Methods}

As shown in Equation~\ref{eq:class_weighted_ddpm_loss}, DDPM minimizes the sum of $KL$ divergence for each class weighted by class proportion $w_c$. We implemented a weighted loss approach using weights inversely proportional to $w_c$. Our method of DiffROP outperforms this naive method by a large margin shown in Table~\ref{tab:reweighted_compare}.

\begin{table}[htb!]
  \centering
  \vspace{-3mm}
  \caption{Performance of DiffROP and Reweighting Methods on CIFAR10LT. Reweighting losses doesn't help while our method improves DDPM.}
  \label{tab:reweighted_compare}
  \begin{tabular}{lll}
    \toprule
    Dataset & Model & \textbf{FID}$\downarrow$ \\
    \midrule
    CIFAR10LT & DDPM~\cite{ho2020denoising} &5.76 	 \\
    \cline{2-3}
    & ~{+DiffROP} &\textbf{5.34}   \\
    & ~{+Reweighting} &6.45   \\
    \bottomrule
  \end{tabular}
\label{tab:weighted_loss}
\end{table}

\subsection{Ablation study}
\label{sec:ablation}

\textbf{Ablation study of different loss types for $\gL_\text{PCL}$} In Table~\ref{tab:loss_type}, our study conducts a comparative analysis of various Probabilistic Contrastive Learning loss functions, including the negative $L_2$ distance and Hinge-like losses in both exponential and reciprocal forms. For comprehensive formulations of these losses, readers are referred to Section \ref{sec:overall_framework}. The initial implementation employing the negative $L_2$ distance yielded suboptimal Fr\'echet Inception Distance (FID) scores, particularly when contrasted with the baseline DDPM model. Consequently, this underscores the imperative of integrating Hinge-like losses, such as those based on exponential and reciprocal functions, to enhance the efficacy of the DDPM framework.

\begin{table}[t!]
  \centering
  \caption{Results of different loss types for $\gL_\text{PCL}$.}
  \label{tab:loss_type}
  \scalebox{0.85}{\begin{tabular}{lll}
    \toprule
    Dataset  &Model 
    & \textbf{FID}$\downarrow$ \\
    \midrule
    CIFAR10LT & DDPM~\cite{ho2020denoising} 
    &5.76 	 \\
    & ~{+DiffROP (Negative $L_2$ Distance)} 
    &11.73   \\
    & ~{+DiffROP (Negative Exponential Form)} 
    &\textbf{5.43}   \\
    & ~{+DiffROP (Reciprocal Form)} 
    &5.72  \\
    \midrule
    CIFAR100LT & DDPM~\cite{ho2020denoising} 
    &7.38 	 \\
    & ~{+DiffROP (Negative $L_2$ Distance)} 
    &12.43  \\
    & ~{+DiffROP (Negative Exponential Form)} 
    &\textbf{6.84}   \\
    & ~{+DiffROP (Reciprocal Form)} 
    &6.98   \\
    \bottomrule
  \end{tabular}}
\end{table}

\begin{figure}[!ht]
\centering
\subfigure[\textit{Ablation study}]{
\begin{minipage}[t]{0.5\linewidth}
\centering
\includegraphics[scale=0.097]{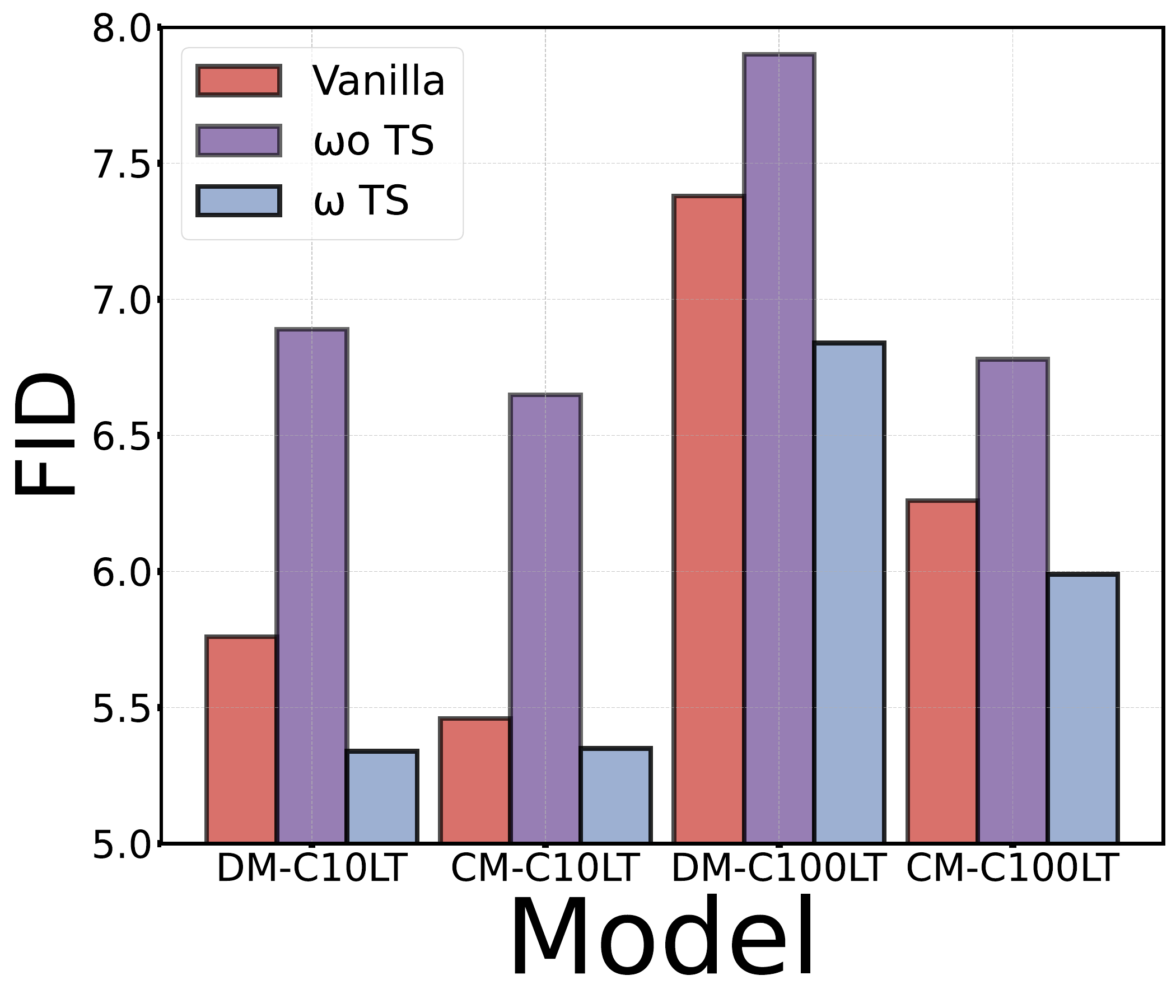}
\label{ablation-t}
\end{minipage}%
}%
\subfigure[\textit{Sensitivity Analysis}]{
\begin{minipage}[t]{0.5\linewidth}
\centering
\includegraphics[scale=0.2]{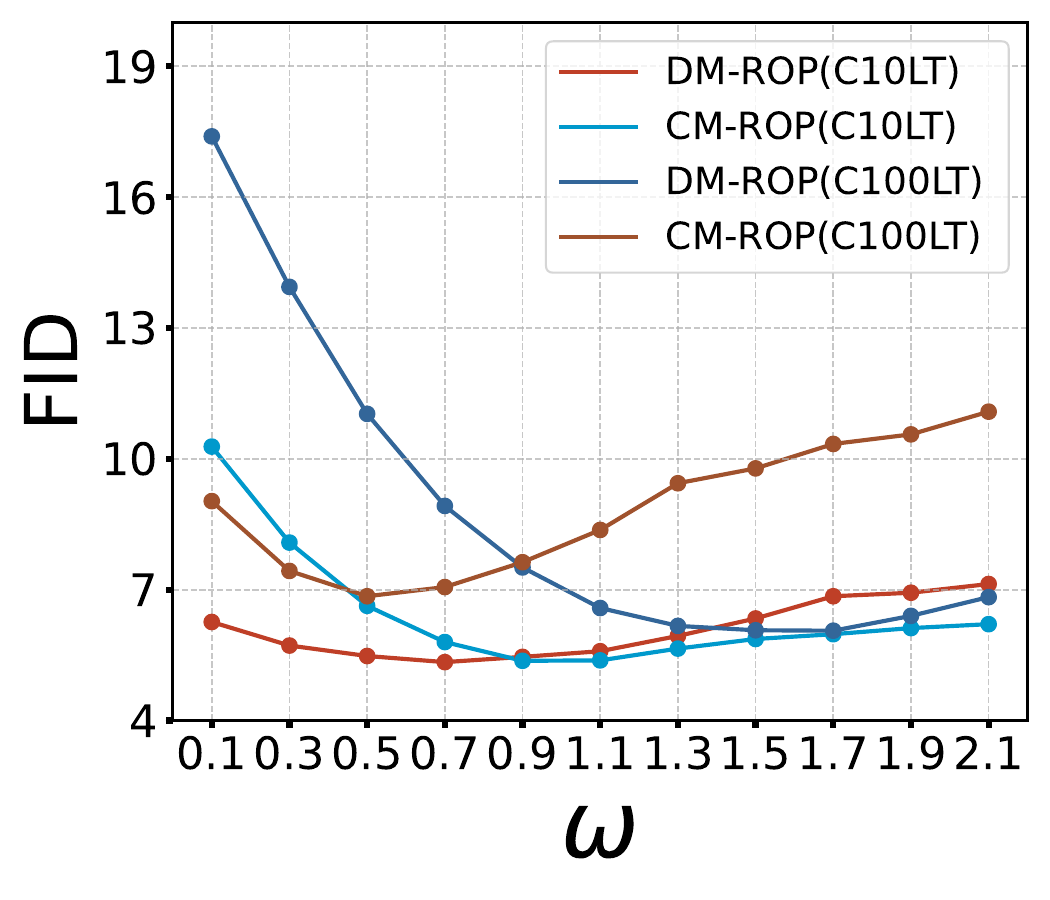}
\label{omega_analysis}
\end{minipage}%
}%
\centering
\caption{ a). We conducted a thorough analysis of the impact that varying the time-dependent parameter, $\tau$, has on the CIFAR10LT and CIFAR100LT datasets. This analysis was performed through a detailed ablation study. 
`TS' refers to the DiffROP model with time-dependent $\tau$.
The term `Vanilla' signifies the standard Vanilla DDPM.
b). We demonstrate the impact of the strength of classifier-free guidance, denoted as $\omega$, on the efficacy of the DiffROP sampling process.}
\vspace{-5mm}
\end{figure}

\paragraph{Ablation study of time-dependent $\tau$.}
Figure~\ref{ablation-t} presents the outcomes of the ablation study focusing on time-dependent $\tau$. The illustrative data in this figure highlight the significant contribution of time-dependent $\tau$ to improving DiffROP's efficacy, particularly with long-tailed datasets. A notable enhancement in the FID scores is observed when comparing the performances against the $\text{Vanilla}$ and non-time-dependent configurations. The empirical evidence suggests that the judicious choice of a $\tau$ that corresponds appropriately to each time step $T$ is instrumental in optimizing the performance of DiffROP.

\paragraph{Hyperparameter sensitivity analysis of $\omega$ in Classifier-free Guidance}
In Figure \ref{omega_analysis}, the performance evaluation of DiffROP is illustrated through FID visualization across a spectrum of $\omega$ values. The graphic representation vividly demonstrates the varying effect of classifier-free guidance strength $\omega$ on the model's overall performance. Our hypothesis suggests that elevated $\omega$ values have a detrimental effect on DiffROP's effectiveness. Conversely, opting for smaller values of $\omega$ holds promise for notably augmenting DiffROP's performance metrics. This observation underscores the importance of fine-tuning $\omega$ parameters to optimize DiffROP's efficacy in diverse contexts.

\vspace{-3mm}
\section{Conclusion}
\vspace{-1mm}
Deep generative models, including diffusion models, are biased towards classes with abundant training images. We introduce DiffROP, a novel framework for training class-imbalanced diffusion models. We are motivated by the observed appearance overlap between head class images and tail class images. We propose probabilistic contrastive learning losses to minimize distribution overlaps for different classes. Our loss is modular, simple to implement, and adds little computational overhead during training. Variants of our probabilistic contrastive learning methods are shown effective in extensive experiments on multiple datasets with long-tailed distributions.

\vspace{-2mm}
\paragraph{Broader Impact} Our research begins with addressing the long-tailed distribution problem in generative models and simultaneously aims to tackle the ethical issue of life imbalance in an inherently unequal world. We recognize the potential for unforeseen challenges, but our future objective is to address long-tail distributions more broadly, encompassing aspects like sex, culture, and race. We aspire for this work to encourage more extensive research in this area.


\clearpage

\bibliography{ref}
\bibliographystyle{icml2024}

\newpage
\appendix
\onecolumn

\section{Proof of Proposition \ref{prop:weighted_kl}}

\label{proof_by_class}
We denote the number of all training images as $N$, and the number of training images for class $\mathbf{c}^i\in \mathcal{C}$ as $N^i$. Following \citet{ho2020denoising},  the training objective of DDMP in Equation~\ref{eq:kl_loss} can be rewritten as
\begin{equation}
    \begin{aligned}
        &\mathbb{E}_{q}\Biggl\{\sum_{t\geq1}^{T}D_{KL}\left[q(x_{t-1}|x_{t},x_{0},\mathbf{c})||p_{\theta}(x_{t-1}|x_{t},\mathbf{c})\right]\Biggr\}\\
        &=\sum_{t\geq1}^{T}\mathbb{E}_{q}\Biggl\{D_{KL}\left[q(x_{t-1}|x_{t},x_{0},\mathbf{c})||p_{\theta}(x_{t-1}|x_{t},\mathbf{c})\right]\Biggr\}\\
        &=\sum_{t\geq1}^{T}\Biggl\{\frac{1}{N}\sum_{(x_0,\mathbf{c})}D_{KL}\left[q(x_{t-1}|x_{t},x_{0},\mathbf{c})||p_{\theta}(x_{t-1}|x_{t},\mathbf{c})\right]\Biggr\}\\
        &=\sum_{t\geq1}^{T}\Biggl\{\frac{1}{N} \sum_{\mathbf{c}^i\in\mathcal{C}}\sum_{(x_0,\mathbf{c}^i)}D_{KL}\left[q(x_{t-1}|x_{t},x_{0},\mathbf{c}^i)||p_{\theta}(x_{t-1}|x_{t},\mathbf{c}^i)\right]\Biggr\}\\
        &=\sum_{t\geq1}^{T}\Biggl\{\frac{1}{N} \sum_{\mathbf{c}^i\in\mathcal{C}} N^i \mathbb{E}_{q,\mathbf{c}^i}\Biggl\{D_{KL}\left[q(x_{t-1}|x_{t},x_{0},\mathbf{c}^i)||p_{\theta}(x_{t-1}|x_{t},\mathbf{c}^i)\right]\Biggr\}\Biggr\}\\
        &=\sum_{t\geq1}^{T}\Biggl\{\sum_{\mathbf{c}^i\in\mathcal{C}}  \frac{N^i}{N} \mathbb{E}_{q,\mathbf{c}^i}\Biggl\{D_{KL}\left[q(x_{t-1}|x_{t},x_{0},\mathbf{c}^i)||p_{\theta}(x_{t-1}|x_{t},\mathbf{c}^i)\right]\Biggr\}\Biggr\}\\
        &=\sum_{\mathbf{c}^i\in\mathcal{C}}  \frac{N^i}{N} \mathbb{E}_{q,\mathbf{c}^i} \Biggl\{\sum_{t\geq1}^{T} \Biggl\{D_{KL}\left[q(x_{t-1}|x_{t},x_{0},\mathbf{c}^i)||p_{\theta}(x_{t-1}|x_{t},\mathbf{c}^i)\right]\Biggr\}\Biggr\}\\
        &= \sum_{\mathbf{c}^i\in\mathcal{C}} \underset{\text{Learning Bias}}{\underline{{w_i}}} \mathbb{E}_{q,\mathbf{c}^i} \Biggl\{\sum_{t\geq1}^{T} D_{KL}\left[q(x_{t-1}|x_{t},x_{0},\mathbf{c}^i)||p_{\theta}(x_{t-1}|x_{t},\mathbf{c}^i)\right]\Biggr\},\\
    \end{aligned}
\end{equation}

where $w_i=\frac{N^i}{N}$ is the ratio of class $\mathbf{c}^i$ images in the training set.

\section{Visualization of Images Generated under the Condition of Tail and Random Classes}
\begin{figure*}
\centering
\begin{tabular}{cc}
    \includegraphics[width=0.45\columnwidth,trim={0 7.2cm 0 0},clip]{cifar10lt_ddpm_300000_ddpm_baseline_N144_STEP300000_new_low.pdf} &
    \includegraphics[width=0.45\columnwidth,trim={0 7.2cm 0 0},clip]{cifar10lt_cbdm_300000_cbdm_baseline_N144_STEP300000_new_low.pdf} \\
    \small{(a) DDPM} & (b) CBDM \\
    \includegraphics[width=0.45\columnwidth,trim={0 7.2cm 0 0},clip]{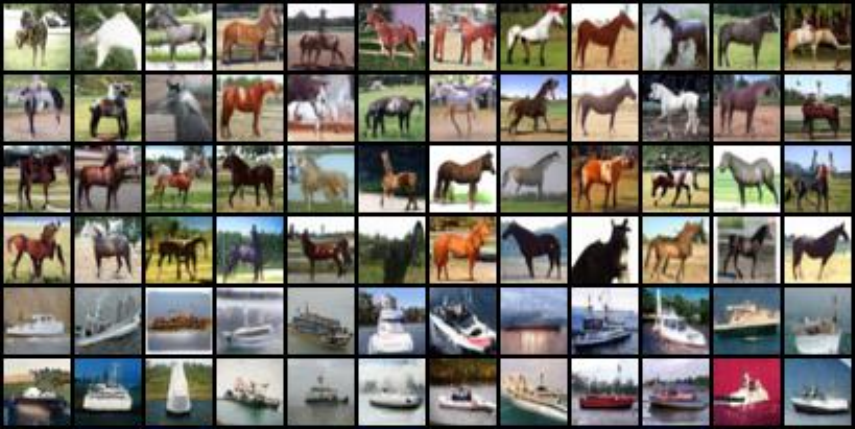} &
    \includegraphics[width=0.45\columnwidth,trim={0 7.2cm 0 0},clip]{cifar10lt_cbdm_300003_divin_N144_STEP300003_new_low.pdf} \\
    (c) DDPM+DiffROP & (d) CBDM+DiffROP \\
\end{tabular}
\caption{Qualitative results for tail classes in CIFAR10LT. Our method is better at creating clear and realistic images for less common classes compared to other basic methods.}
\end{figure*}
\begin{figure*}
\centering
\begin{tabular}{cc}
    \includegraphics[width=0.45\columnwidth,trim={0 7.2cm 0 0},clip]{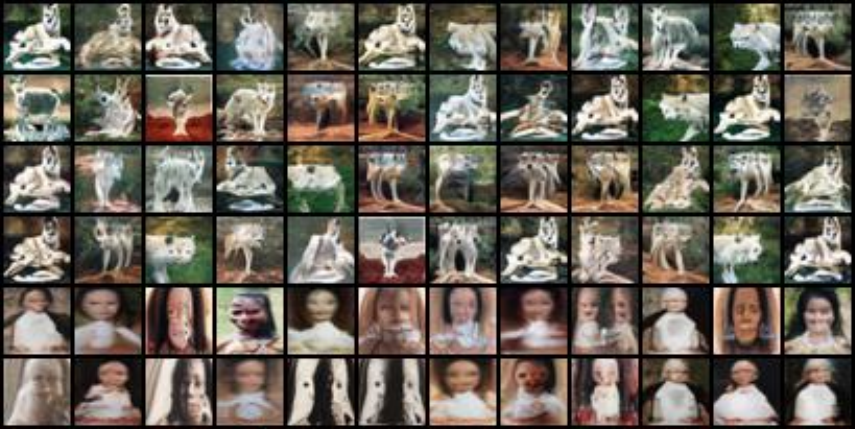} &
    \includegraphics[width=0.45\columnwidth,trim={0 7.2cm 0 0},clip]{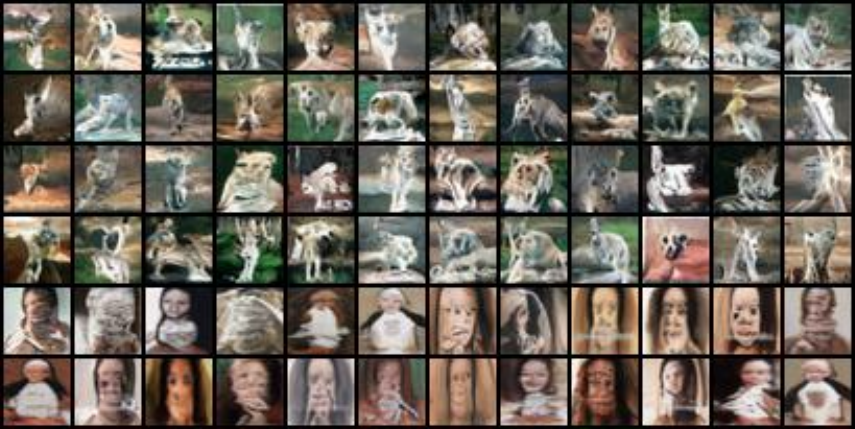} \\
    \small{(a) DDPM} & (b) CBDM \\
    \includegraphics[width=0.45\columnwidth,trim={0 7.2cm 0 0},clip]{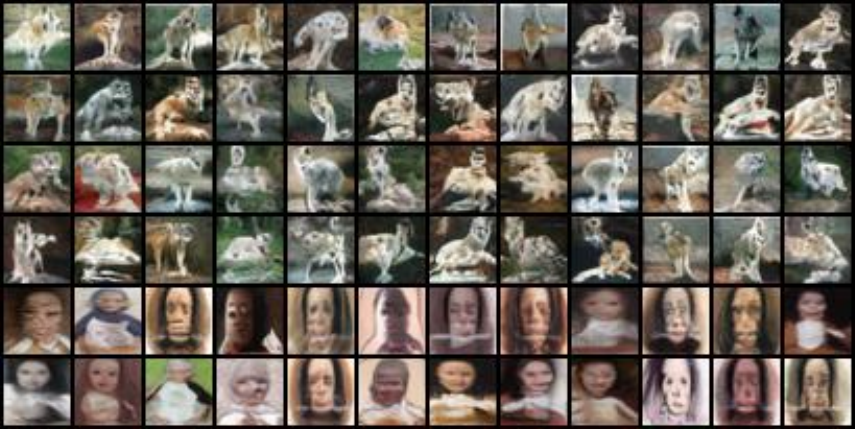} &
    \includegraphics[width=0.45\columnwidth,trim={0 7.2cm 0 0},clip]{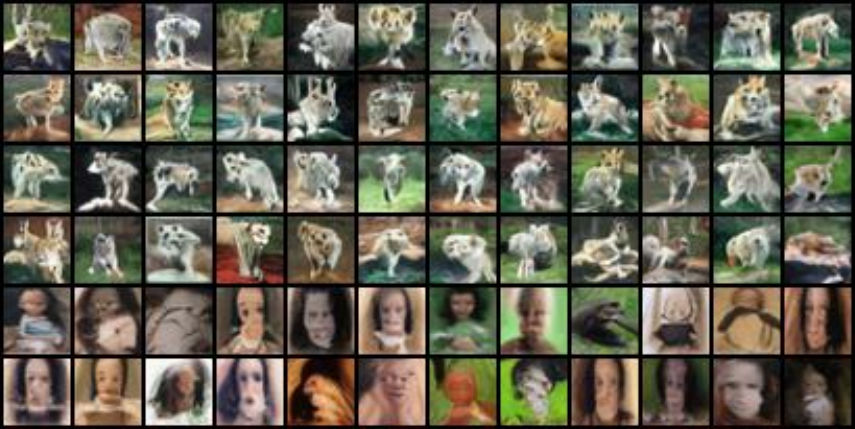} \\
    (c) DDPM+DiffROP & (d) CBDM+DiffROP \\
\end{tabular}
\caption{Qualitative results for tail classes in CIFAR100LT. Our method is better at creating clear and realistic images for less common classes compared to other basic methods.}
\end{figure*}
\begin{figure*}
\centering
\begin{tabular}{cc}
    \includegraphics[width=0.45\columnwidth,trim={0 7.2cm 0 0},clip]{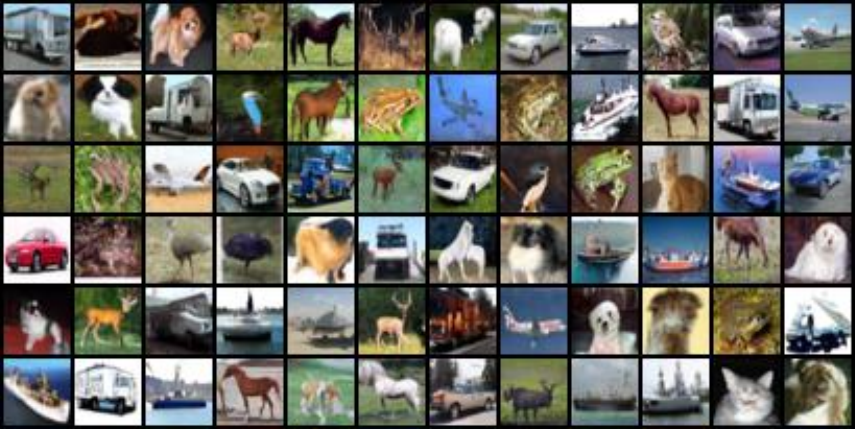} &
    \includegraphics[width=0.45\columnwidth,trim={0 7.2cm 0 0},clip]{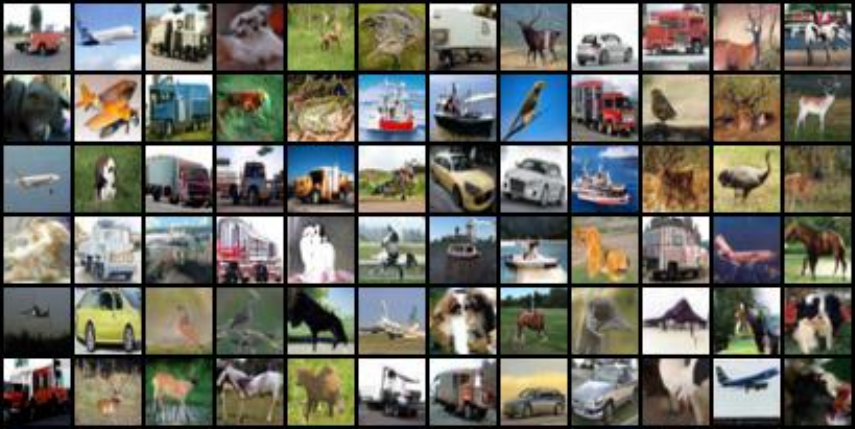} \\
    \small{(a) DDPM} & (b) CBDM \\
    \includegraphics[width=0.45\columnwidth,trim={0 7.2cm 0 0},clip]{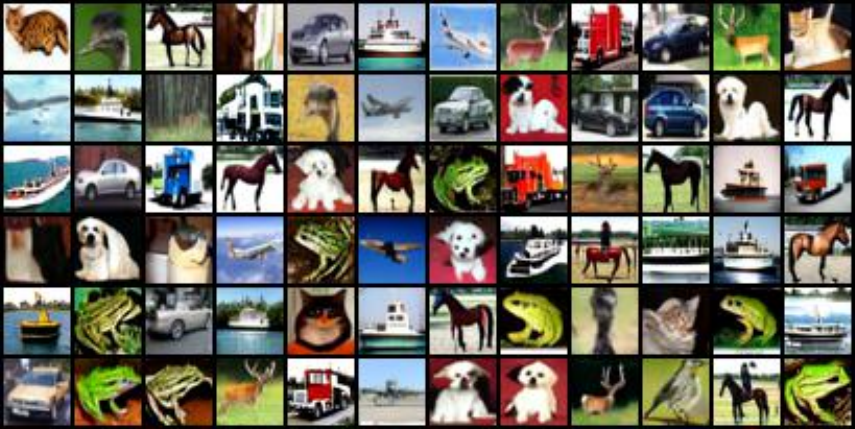} &
    \includegraphics[width=0.45\columnwidth,trim={0 7.2cm 0 0},clip]{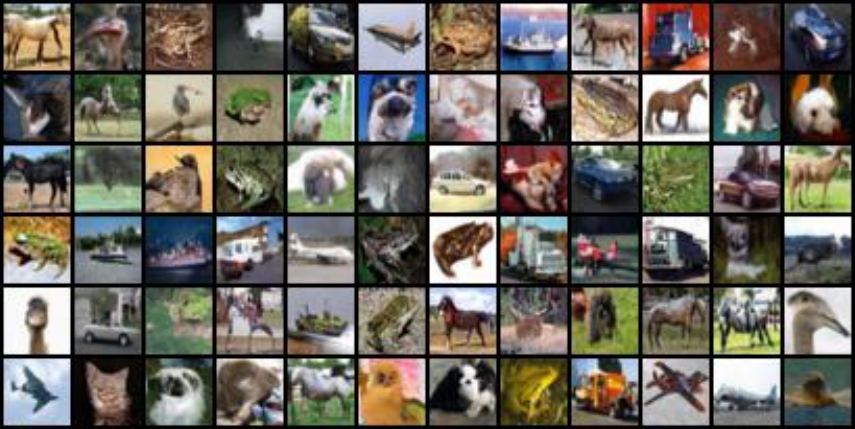} \\
    (c) DDPM+DiffROP & (d) CBDM+DiffROP \\
\end{tabular}
\caption{Qualitative results for random classes in CIFAR10LT. Our method is better at creating clear and realistic images for most classes compared to other basic methods.}
\label{fig:random_class_images}
\end{figure*}
\begin{figure*}
\centering
\begin{tabular}{cc}
    \includegraphics[width=0.45\columnwidth,trim={0 7.2cm 0 0},clip]{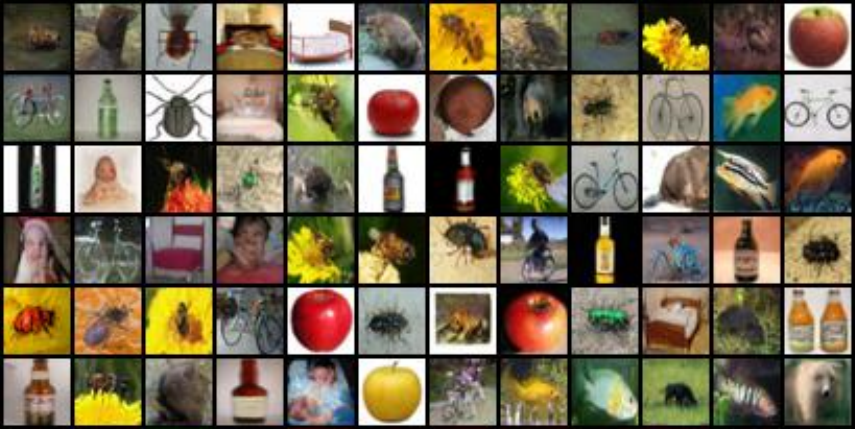} &
    \includegraphics[width=0.45\columnwidth,trim={0 7.2cm 0 0},clip]{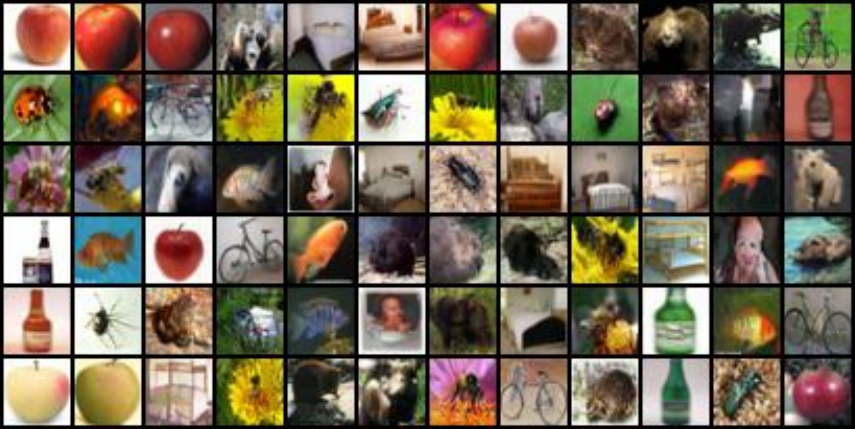} \\
    \small{(a) DDPM} & (b) CBDM \\
    \includegraphics[width=0.45\columnwidth,trim={0 7.2cm 0 0},clip]{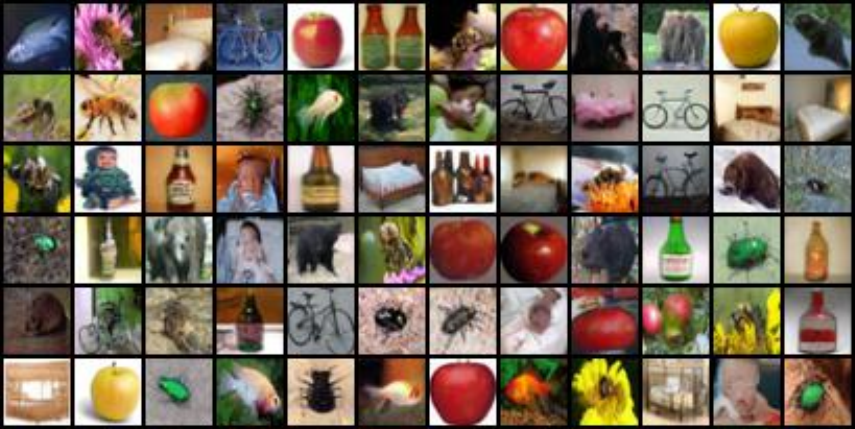} &
    \includegraphics[width=0.45\columnwidth,trim={0 7.2cm 0 0},clip]{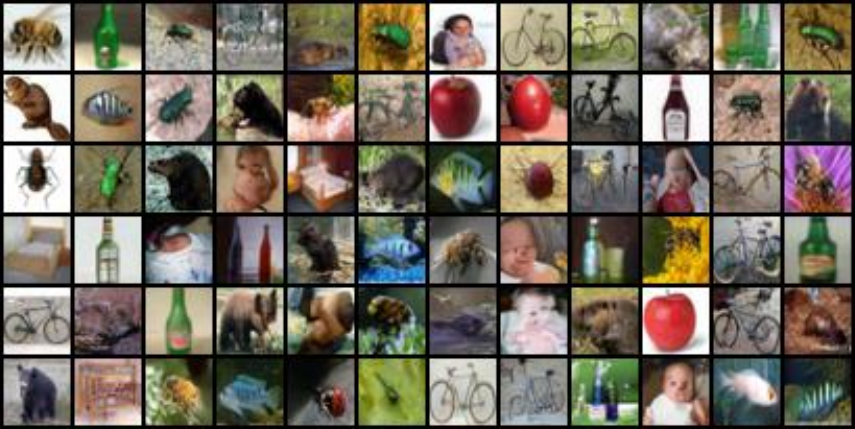} \\
    (c) DDPM+DiffROP & (d) CBDM+DiffROP \\
\end{tabular}
\caption{Qualitative results for random classes in CIFAR100LT. Our method is better at creating clear and realistic images for most classes compared to other basic methods.}
\end{figure*}

\end{document}

%% file: math_commands.tex
\usepackage{amsmath,amsfonts,bm}

\def\eqref#1{equation~\ref{#1}}
\def\Eqref#1{Equation~\ref{#1}}

\def\1{\bm{1}}

\def\rvepsilon{{\mathbf{\epsilon}}}

\def\rvx{{\mathbf{x}}}

\def\vzero{{\bm{0}}}

\def\mI{{\bm{I}}}

\DeclareMathAlphabet{\mathsfit}{\encodingdefault}{\sfdefault}{m}{sl}
\SetMathAlphabet{\mathsfit}{bold}{\encodingdefault}{\sfdefault}{bx}{n}

\def\gL{{\mathcal{L}}}

\def\gN{{\mathcal{N}}}

\newcommand{\E}{\mathbb{E}}

\newcommand{\KL}{D_{\mathrm{KL}}}

